\documentclass[12pt,twoside]{article}
\usepackage[super,sort,comma]{natbib}
\usepackage{ulem}
\usepackage{amsmath}
\usepackage{amssymb}
\usepackage{fancyhdr}

\usepackage{graphicx}
\makeatletter \renewcommand\@biblabel[1]{$^{#1}$} \makeatother
\newcommand{\note}[1]{\mbox{}\\ \noindent \rule{16cm}{0.5mm} \\
{\em #1} \\ \noindent \rule{16cm}{0.5mm}
\typeout{    }
\typeout{***********note active on this page *************************}
\typeout{Note: #1  }
\typeout{****************************************end Note}
}
\renewcommand{\note}[1]{}

\newcommand{\cen}[1]{\begin{center} #1 \end{center}}

\usepackage[mathlines]{lineno}

\usepackage[section]{placeins}
\usepackage{hyperref}
\hypersetup{ colorlinks,
    citecolor=blue,
    filecolor=blue,
    linkcolor=blue,
    urlcolor=blue
}
\usepackage{xcolor}
\usepackage[all]{hypcap} 
\definecolor{gray}{rgb}{0.6,0.6,0.6}
\definecolor{red}{rgb}{0.85,0,0}
\definecolor{green}{rgb}{0,0.85,0}
\definecolor{blue}{rgb}{0,0,0.85}
\definecolor{beige}{rgb}{0.92,0.87,0.78}

\begin{document}

\cen{\sf {\Large {\bfseries Multi-Granularity Vision Fastformer with Fusion Mechanism for Skin Lesion Segmentation } \\  
\vspace*{10mm}
Xuanyu Liu$^{a}$, Huiyun Yao$^{a}$, Jinggui Gao$^{a}$, Zhongyi Guo$^{a}$, Xue Zhang$^{a}$, Yulin Dong$^{a}$} \\
$^{a}$College of Mathematical and Systems Science, Shandong University of Science and Technology, Huangdao, Qingdao, Shandong 266590, China
\vspace{5mm}\\
}

\pagenumbering{roman}
\setcounter{page}{1}
\pagestyle{plain}
Corresponding author: Jinggui Gao, email: jingguigao@126.com\\

\begin{abstract}
\noindent {\bf Background:}  Convolutional Neural Networks(CNN) and Vision Transformers(ViT) are the main techniques used in Medical image segmentation. However, CNN is limited to local contextual information, and ViT's quadratic complexity results in significant computational costs. At the same time, equipping the model to distinguish lesion boundaries with varying degrees of severity is also a challenge encountered in skin lesion segmentation.\\ 
{\bf Purpose:} This research aims to optimize the balance between computational costs and long-range dependency modelling and achieve excellent generalization across lesions with different degrees of severity.\\
{\bf Methods:} we propose a lightweight U-shape network that utilizes Vision Fastformer with Fusion Mechanism (VFFM-UNet). We inherit the advantages of Fastformer's additive attention mechanism, combining element-wise product and matrix product for comprehensive feature extraction and channel reduction to save computational costs. In order to accurately identify the lesion boundaries with varying degrees of severity, we designed Fusion Mechanism including Multi-Granularity Fusion and Channel Fusion, which can process the feature maps in the granularity and channel levels to obtain different contextual information.\\
{\bf Results:}Comprehensive experiments on the ISIC2017, ISIC2018 and PH$^{2}$ datasets demonstrate that VFFM-UNet outperforms existing state-of-the-art models regarding parameter numbers, computational complexity and segmentation performance. In short, compared to MISSFormer, our model achieves superior segmentation performance while reducing parameter and computation costs by 101x and 15x, respectively. \\
{\bf Conclusions:} Both quantitative and qualitative analyses show that VFFM-UNet sets a new benchmark by reaching an ideal balance between parameter numbers, computational complexity, and segmentation performance compared to existing state-of-the-art models. \\

\end{abstract}
\noindent {\bf Keywords:} Medical image segmentation, Fastformer, Fusion Mechanism.
\newpage     

\newpage

\setlength{\baselineskip}{0.7cm}      

\pagenumbering{arabic}
\setcounter{page}{1}
\pagestyle{fancy}
\section{Introduction}
A report from the American Society of Clinical Oncology (ASCO) reveals that malignant melanoma is increasing rapidly, making it one of the fastest-growing tumour types. In the last decade, there have been approximately 160,000 new cases and 48,000 deaths per year worldwide. Because of this, there is an urgent need for automated skin lesion segmentation systems to assist medical professionals in quickly and accurately identifying the areas of the lesion. In the field of medical image segmentation, Convolutional Neural Networks (CNN) and Vision Transformers (ViT) \cite{19} are the main applied techniques. However, both techniques have limitations: the perspective of CNN network models is limited to local contextual information, and they are almost unable to model global long-range dependencies; At the same time, ViT can effectively extract global contextual information, but its quadratic complexity results in a significant computational cost. Although some studies focus on exploring more efficient attention mechanisms \cite{8, 12, 13} or constructing lightweight models \cite{9, 10, 11} capable of capturing contextual information, models still need to be deployed in real-world settings, where computational demands, especially in resource-constrained environments, continue to pose challenges \cite{35, 50}. As such, the balance between computational costs and long-range dependency modelling can still be optimized. Additionally, challenges still remain in skin lesion segmentation, such as the difficulty in processing images with extremely low contrast \cite{23}. In this paper, we focus on the issue of identifying unclear lesion boundaries, especially in samples with subtle colour changes, and further aim to achieve excellent generalization across lesions with different degrees of severity.\par
\begin{figure*}[t]
\centering
\includegraphics[width=1\textwidth]{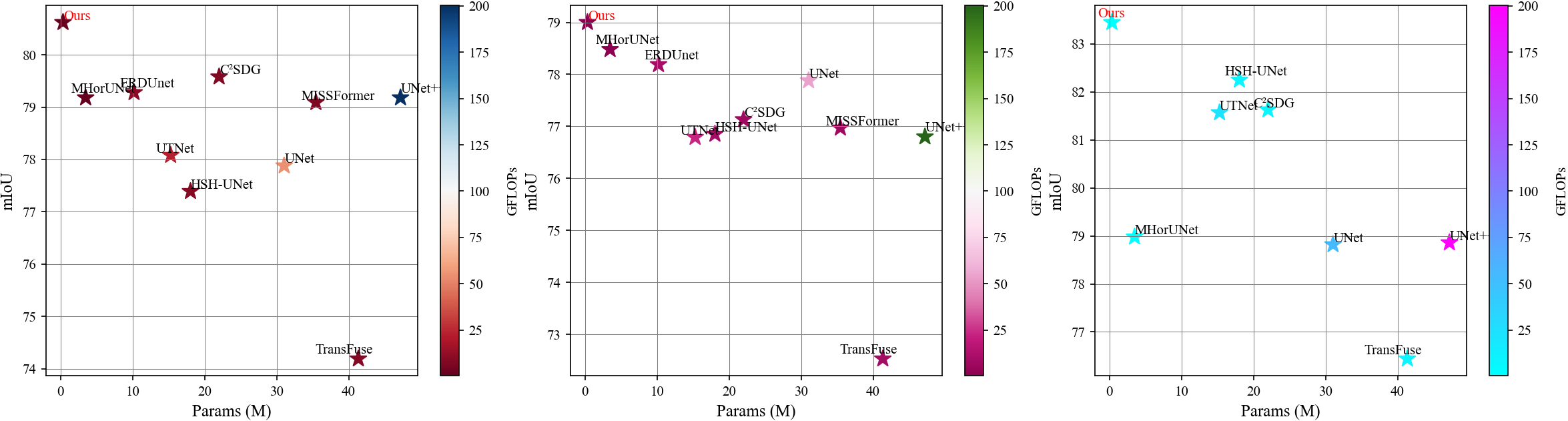}
\caption{From left to right, the visualizations show the comparative experimental results on the ISIC2018, ISIC2017, and PH2 datasets. The X-axis represents the number of parameters (lower is better), while Y-axis represents mIoU (higher is better). The color depth represents computational complexity (GFLOPs, lighter is better).}
\label{fig7}
\end{figure*}
In recent years, Fastformer \cite{14}, an efficient Transformer variant, has shown strong performance in Natural Language Processing(NLP). At the same time, several studies \cite{15, 16, 17} have shown that Fastformer performs well in many fields. On the one hand, Fastformer, based on the additive attention mechanism, can achieve powerful feature extraction with linear complexity. On the other hand, the model learns global context-aware attention values through the interaction between the global query and key vectors, which enables it to finish global long-range dependency modelling. Inspired by Fastformer,  Fast Vision Transformer \cite{18} is the first to introduce Fastformer into the visual domain, achieving remarkable results in image classification. However, many precedents have proven that applying language models to visual tasks requires adapting how sequence data is processed to accommodate image data. Vision Transformer and Vision Mamba \cite{20} are two typical examples. To facilitate the establishment of global receptive fields in the 2D space, Vision Transformer designs Patch Embedding and Positional Encoding. At the same time, Vision Mamba incorporates the Cross-Scan module into the Selective Scan Mechanism. In the experimental section of our paper, we attempted to apply the vanilla Fastformer to our task, but the results are unsatisfactory. We deduce that the issue arises from the fact that relying solely on the element-wise product for feature extraction leads to insufficient feature representation. Therefore, some adjustments are necessary to better adapt the model to our task. In summary, we aim to leverage the inherent advantages of Fastformer in visual tasks and conduct a deeper exploration of it by optimizing the way the model processes image data to fully harness its potential in the field of skin lesion segmentation.\par
In order to gain a better understanding of the key aspects of solving the challenges mentioned above, it's essential to conduct a thorough and detailed analysis of the datasets about skin lesion segmentation. Although several studies \cite{21, 22, 1, 2, 4, 23, 24, 51, 52} have achieved promising results in this field, relatively few have developed models that account for the varying degrees of severity. According to an analysis of the skin lesion data, We can draw two insightful conclusions:\par

\begin{figure}[t]
\centering
\includegraphics[width=0.46\textwidth]{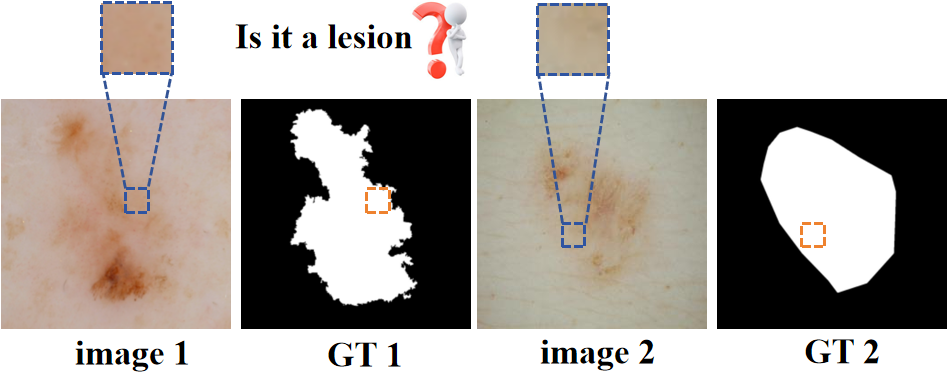}
\caption{Accurately identifying the unclear boundaries of lesions requires certain contextual information.}
\label{fig9}
\end{figure}

\noindent
\hangindent=1.2em  
\hangafter=1     
\noindent
1. \textbf{Accurately identifying the unclear boundaries of lesions requires certain contextual information.} As shown in Fig. \ref{fig9}, we can observe that boundaries with a mild degree of severity often have subtle colour changes, making them similar to normal skin and difficult to distinguish. However, if we take a global perspective and have more contextual information, the likelihood of misjudgment is significantly reduced.\par
\begin{figure}[t]
\centering
\includegraphics[width=0.46\textwidth]{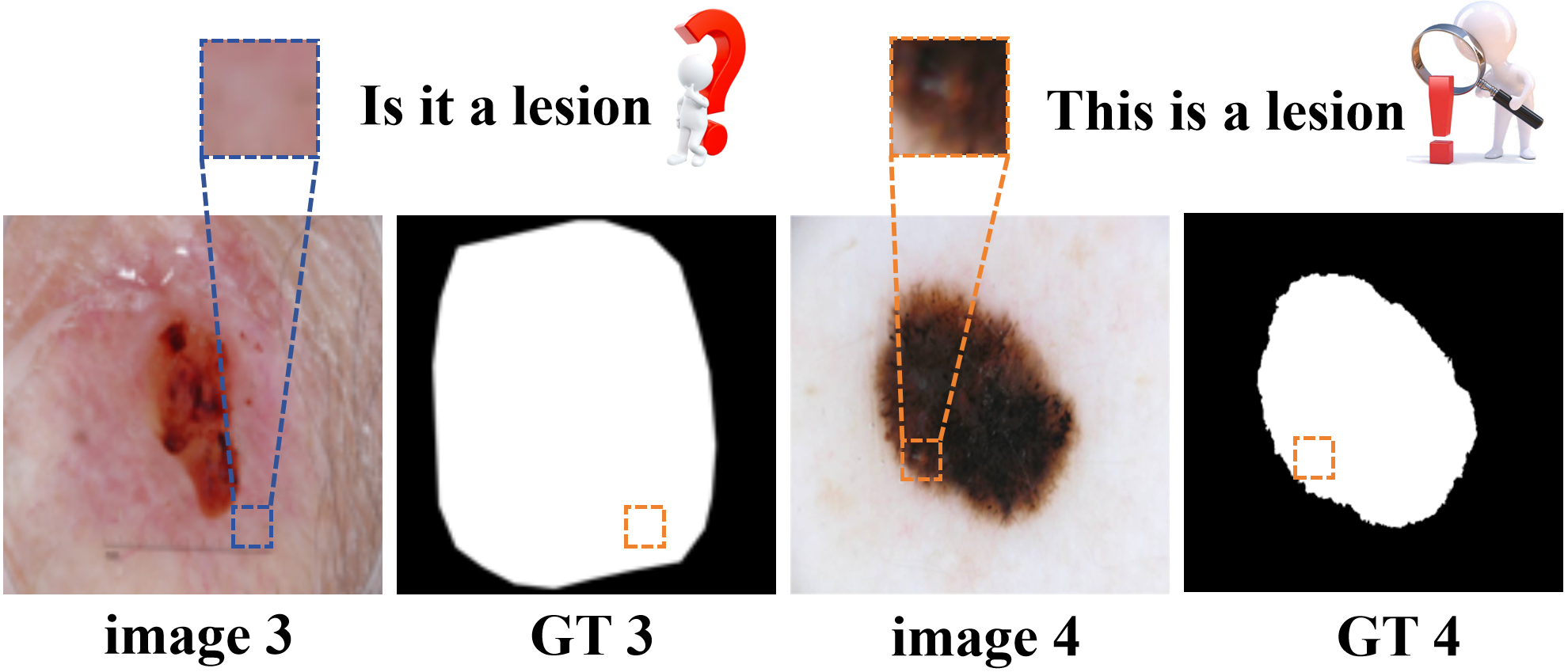}
\caption{Lesions with different degrees of severity require different contextual information for identification of their boundaries.}
\label{fig10}
\end{figure}

\noindent
\hangindent=1.2em  
\hangafter=1 
\noindent
2. \textbf{Lesions with different degrees of severity require different contextual information to identify their boundaries.} As shown in Fig. \ref{fig10}, compared to the lesion in Image 3, we can clearly identify the lesion boundaries with a severe degree of severity due to the obvious changes in colour in Image 4. Therefore, the contextual information we require is not fixed; it dynamically changes with the skin lesion data of different degrees of severity.\par
To address the aforementioned challenge, which requires a wide and different range of contextual information, we propose VFFM-UNet, built upon the U-shape architecture following a 6-stage encoder-decoder structure. The model's core components are Multi-Granularity Vision Fastformer(MGVF) and Fusion Mechanism(FM). In MGVF, we introduce Vision Fastformer, which achieves a good trade-off between computational costs and long-range dependency modelling. Furthermore, we leverage it to extract feature maps at three different granularities, allowing us to obtain contextual information at different levels. In FM, these feature maps contain different contextual information fused at both the granularity and channel levels. This fusion empowers the model to accurately identify the lesion boundaries with varying degrees of severity, thereby significantly boosting its generalization ability.\par
In summary, our contributions can be categorized into the following three aspects:\par
\noindent
\hangindent=1.5em  
\hangafter=1 
\noindent
$\bullet$\quad We propose VFFM-UNet, a hybrid architecture network. We introduce a language model, Fastformer, into skin lesion segmentation for the first time. By tackling the challenge of identifying unclear lesion boundaries, we explored the potential of the model for this task. The model achieves good performance with 0.35M parameters and 0.494 GFLOPs, effectively balancing computational costs and long-range dependency modelling.\par

\begin{figure*}[t]
    \centering
    \begin{minipage}{0.4\textwidth}  
        \centering
        \includegraphics[width=1\textwidth]{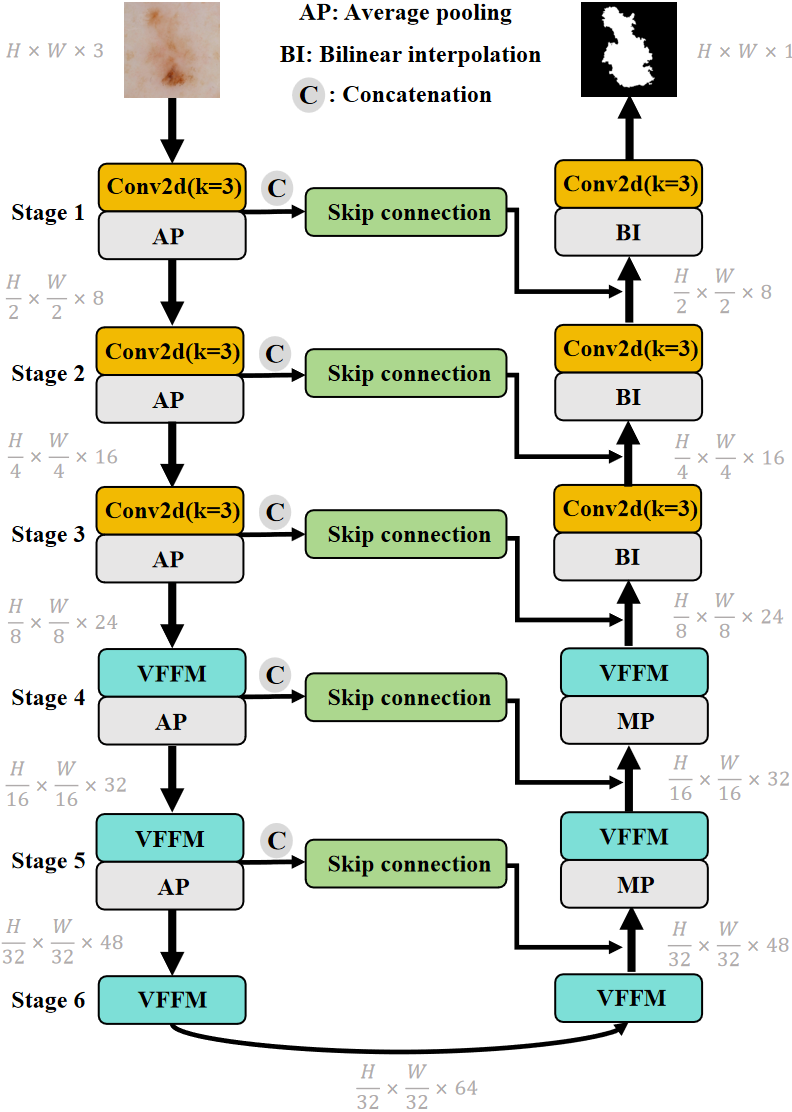}
        \caption{An overview of our proposed VFFM-UNet}
        \label{fig2}
    \end{minipage}
    \hspace{1.2cm}
    \begin{minipage}{0.4\textwidth}
        \centering
        \includegraphics[width=1\textwidth]{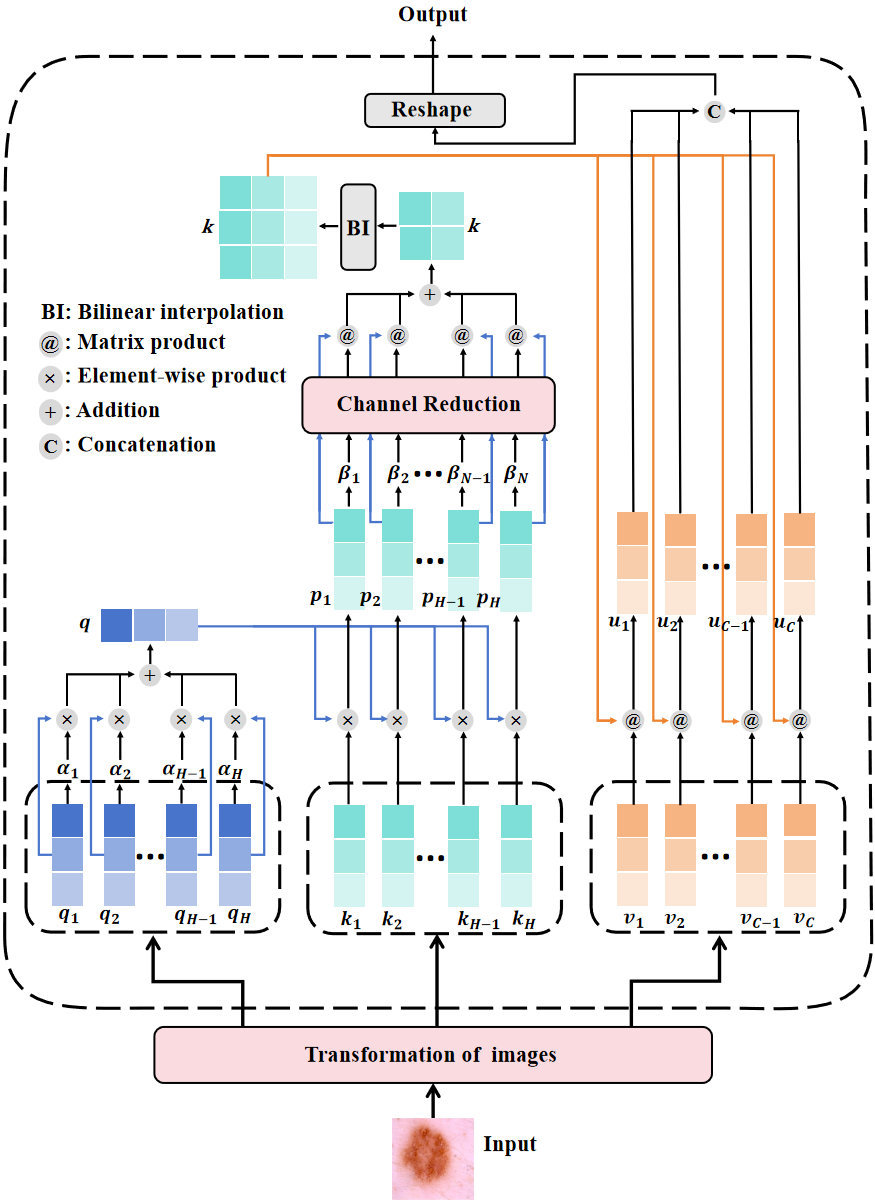}
        \caption{An overview of Vision Fastformer}
        \label{fig1}
    \end{minipage}
\end{figure*}
\noindent
\hangindent=1.5em  
\hangafter=1 
\noindent
$\bullet$\quad We introduce Multi-Granularity Vision Fastformer to extract feature maps at different granularities and incorporate Fusion Mechanism, including Multi-Granularity Fusion and Channel Fusion, to accomplish the model's generalization ability in lesions with a different degree of severity.\par
\noindent
\hangindent=1.5em  
\hangafter=1 
\noindent
$\bullet$\quad Extensive experiments on three datasets for public skin lesion segmentation, ISIC2017, ISIC2018, and PH$^{2}$ dataset, demonstrate that VFFM-UNet is state-of-the-art in terms of number of parameters, computational complexity, and segmentation performance.\par
\section{Related Work}
\subsection{Skin lesion segmentation}
Conventional models for skin lesion segmentation are highly dependent on characteristics such as colour, texture, and others. Taking colour as an example, MEDS \cite{25}, a histogram-based thresholding method, uses a single parameter to achieve the extraction of colour distributions to control how 'tight' the segmentation is. The image threshold method \cite{26} utilizes the Artificial Bee Colony algorithm to choose the optimal threshold values. However, traditional methods often struggle to segment skin lesions accurately, as manually crafted features may not be well-suited for the segmentation task. Therefore, we need methods that can autonomously learn feature extraction.\par
With the development of neural networks, their potential in medical segmentation is gradually being explored. Nowadays, most methods for skin lesion segmentation are based on UNet \cite{6}. Sarker et al. \cite{27} proposed a U-shape network for more accurate segmentation of skin lesion boundaries. To combine the feature maps extracted from the corresponding encoding path and the previous decoding up-convolutional layer in a non-linear way, Azad et al. \cite{28} proposed a Bi-Directional ConvLSTM U-Net. In \cite{11}, UNeXt combined UNet \cite{6} and MLP \cite{30} to achieve a balance between lightweight design and excellent performance. Hu et al. \cite{5} designed a channel-level contrastive single-domain generalization model, where the shallower features of each image and its style-augmented counterpart are extracted and used for contrastive training, resulting in the disentangled style and structure representations. Li et al. \cite{4} proposed an efficient residual double-coding Unet, which includes a CEE module that enables the model to have efficient feature learning ability and a DRA module that can speed up training and optimize segmentation boundary regions by identifying feature region differences across different layers. Wu et al. \cite{1} adopted higher order spatial interaction based on recursive gate convolution and added a multi-stage dimensional fusion mechanism to the skip connection part to form the MHorUNet model architecture with a better generalization capability. These above-mentioned methods have encouraging results, but their large number of parameters and computational complexity make them unsuitable for deployment and practical application in resource-constrained environments.
\begin{figure*}[t]
\centering
\includegraphics[width=0.90\textwidth]{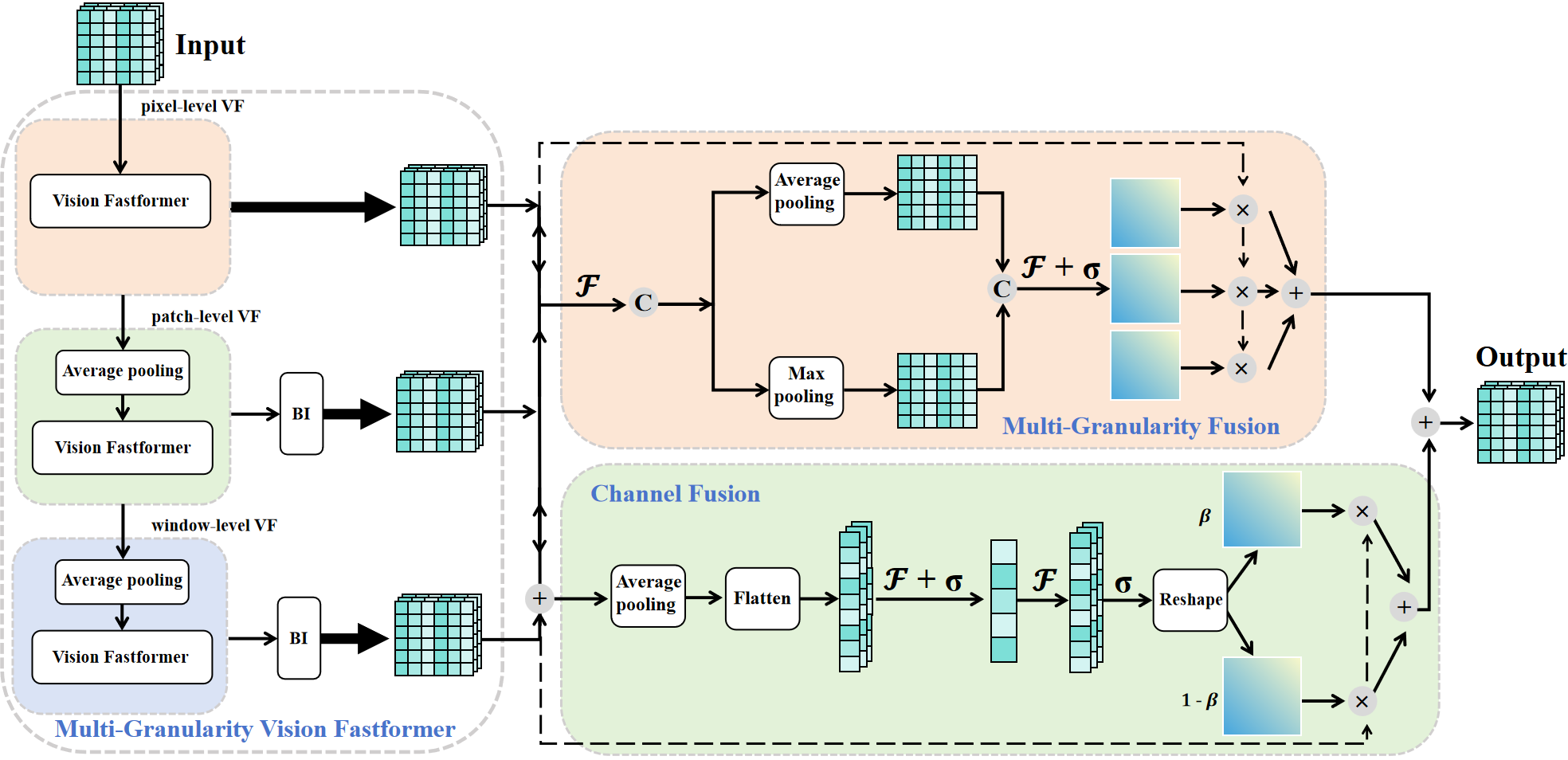}
\caption{An overview of Vision Fastformer with Fusion Mechanism}
\label{fig100}
\end{figure*}
\subsection{Transformers-based UNet model architecture}
The impressive performance of Transformer \cite{31} in Natural Language Processing has sparked interest in its application to computer vision. Recently, Vision Transformer \cite{18} was introduced, and researchers have also explored combining ViT and its variants with UNet model architectures. Chen et al. \cite{32} proposed TransUNet that applies Transformer to the encoder module of UNet to enhance finer details by recovering localized spatial information. To capture local and global contextual information, Zhang et al. \cite{7} employed a dual-path structure applying CNN and ViT simultaneously. Azad et al. \cite{33} innovatively incorporated Transformer into the skip-connections of the standard UNet. They utilized a Spatial Normalization mechanism to adaptively recalibrate the skip connection path, and their methods achieved promising performance. However, Transformer-based models are invariably constrained by their inability to capture local contextual information and typically require large computational resources for training. Therefore, a lightweight model that can model across varying degrees of context information is worth developing.
\section{Method}
In this section, we first introduce the overall architecture of VFFM-UNet, followed by details of the Encoder and Decoder Blocks. Finally, we elaborate on the two core components: Vision Fastformer and Fusion  Mechanism. 
\begin{figure*}[t]
\centering
\includegraphics[width=0.6\textwidth]{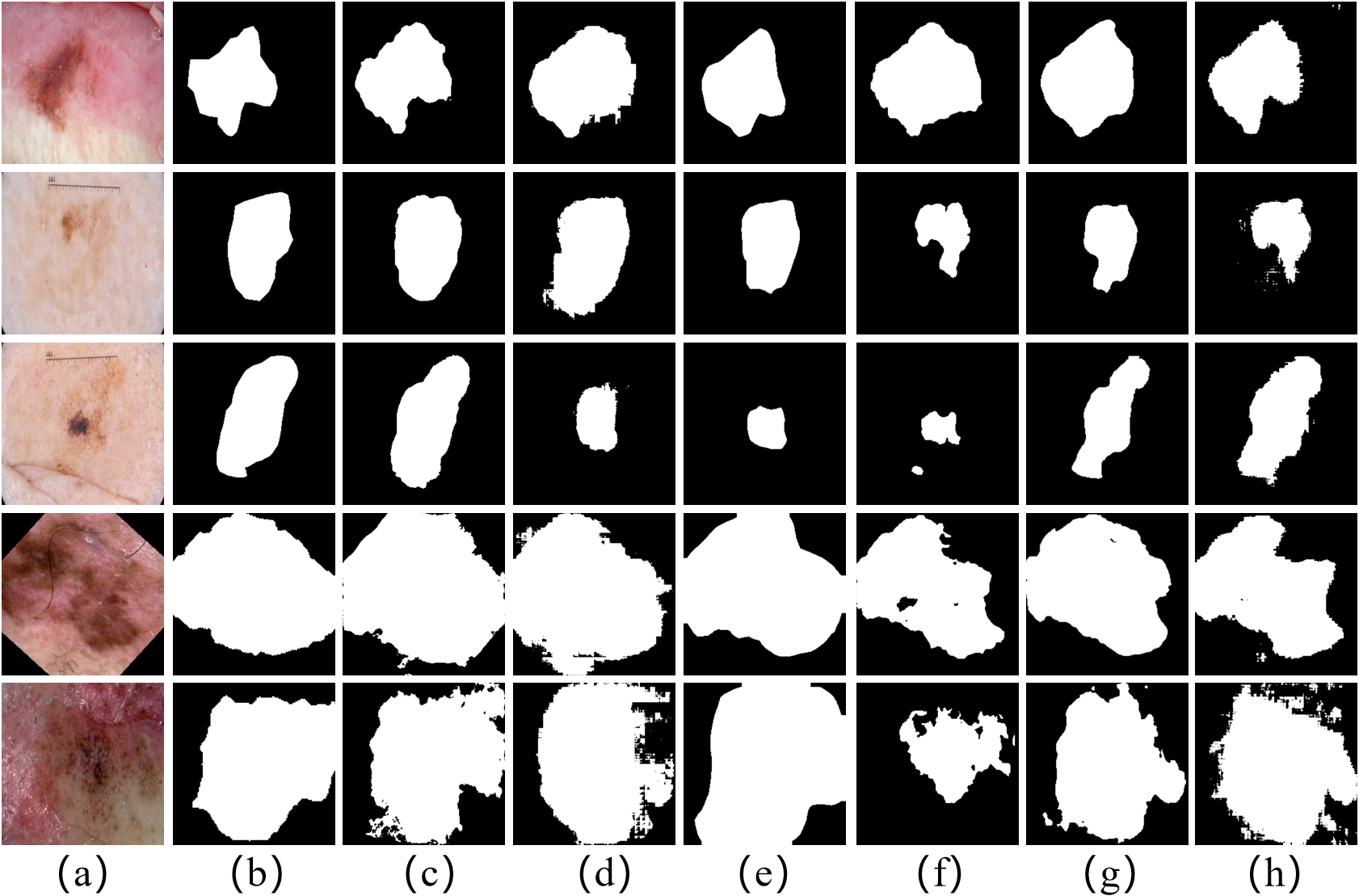}
\caption{Performance of VFFM-UNeISIC 2018. (a) Input images. (b) Groundtruth. The results by (c) VFFM-UNet, (d) ERDUNet, (e) C$^{2}$SDG, (f) MHorunet, (g) HSH-UNet and (h) MISSFormer.}
\label{fig4}
\end{figure*}
\begin{table*}[t]
\centering
\caption{Results for ISIC 2018 Dataset. VFFM-UNet shows significant advantages on segmentation performance.(\textbf{Bold} indicate the best and \uline{underline} indicate the second best.)}
\label{Tab3}
\begin{tabular}{l|c|cc|ccccc}
\hline
Methods & year & Params(M)↓ & GFLOPs↓ & mIoU↑ & DSC↑ & Acc↑ & Sen↑ & Spe↑ \\
\hline
UNet\cite{6} & 2015 & 31.04 & 54.74 & 77.88 & 87.56 & 94.03 & 87.23 & 96.19\\

UNet++\cite{37} & 2018 & 47.19 & 200.12 & 79.18 & 88.38 & 94.40 & 88.27 & 96.34\\

TransFuse\cite{7} & 2021 & 41.34 &  8.87 & 74.19 & 85.18 & 93.02 & 83.27 & 96.12\\

UTNet\cite{38} & 2021 & 15.29 & 22.55 & 78.08 & 87.69 & 93.98 & 89.09 & 95.53 \\

MISSFormer\cite{2} & 2022 & 35.45 & 7.28 & 79.09 & 88.32 & 94.39 & 87.61 & 96.56 \\

C$^{2}$SDG\cite{5} & 2023& 22.01 & 7.97 & \uline{79.58} & \uline{88.63} & 94.43 & 90.13 & 95.79 \\

ERDUnet\cite{4} & 2024& 10.21& 10.29 & 79.28 & 88.44 & 94.38 & \uline{89.03} & 96.09 \\

MHorUNet\cite{1} & 2024& 3.49 & 0.57 & 79.18 & 88.38 & \uline{94.49} & 86.84 & \uline{96.92}\\

HSH-UNet\cite{3} & 2024 & 18.04 & 9.36 & 77.39 & 87.25 & 93.80 & 87.95 & 95.66\\
\hline
Ours & - & 0.35 & 0.494 & \textbf{80.62} & \textbf{89.27} & \textbf{94.73} & \textbf{90.74} & \textbf{97.23} \\
\hline
\end{tabular}
\end{table*}
\subsection{Architecture Overview}
An overview of VFFM-UNet is given in Fig. \ref{fig2}, built upon the U-shape architecture consisting of encoder-decoder parts. VFFM-UNet improves the structure of UNet with the proposed VFFM blocks inserted into the encoder and decoder. First, the feature maps will undergo three stages of standard convolutions, each using a kernel of size 3. Then, the feature maps are fed into the last three stages with the proposed VFFM blocks, and the feature maps fused with different granularities are obtained. Symmetrically, the decoder consists of six stages, with three stages of VFFM blocks first and following three stages of standard convolutions. Between the encoder and decoder, we use the simple skip connection in UNet to fully utilize the feature maps in every stage. A VFFM block contains two parts: Multi-Granularity Vision Fastformer, which has three Vision Fastformer modules, and Fusion Mechanism. For the $l^{th}$ stage containing a VFFM block, the process can be formulated as:
\begin{equation}
\begin{gathered}
\textbf{F}_{l}^{'} = \text{PiVF}(\textbf{F}_{l}), \quad \textbf{F}_{l}^{''} = \text{PaVF}(\textbf{F}_{l}), \quad \textbf{F}_{l}^{'''} = \text{WiVF}(\textbf{F}_{l}), \\
\textbf{F}_{l+1} = \text{FM}(\textbf{F}^{'}, \textbf{F}^{''}, \textbf{F}^{'''})
\end{gathered}
\end{equation}
PiVF, PaVF, and WiVF denote pixel-level VF, patch-level VF, and window-level VF, respectively. When combined with these modules, VFFM-UNet has better segmentation performance and generalization capabilities than previous models.
\subsection{Vision Fastformer}
To address the quadratic complexity issue posed by Transformer, we used Fastformer for the first time in semantic segmentation. The additive attention mechanism proposed by Fastformer summarizes the query and key sequences well into the global query and key vectors. However, after experimentation, we found that the element-wise product results in poor segmentation performance. We speculate that this is due to the significant loss of features during the operations of the image data. Therefore, We propose Vision Fastformer(VF), which combines element-wise product and matrix product for comprehensive feature extraction and uses channel reduction to save computational costs. The architecture of VF is shown in Fig. \ref{fig1}.\par
The input is denoted as \textbf{X} $\in $ $\mathbb{R}^{C\times H\times W}$. The VF first transforms the input into the query, key and value matrix \textbf{Q}, \textbf{K}$\in $ $\mathbb{R}^{Head\times HW\times 1}$, \textbf{V}$\in $ $\mathbb{R}^{C\times HW\times 1}$, which are written as \textbf{Q} = $[q_{1},q_{2},\cdots,q_{H}]$, \textbf{K} = $[k_{1},k_{2},\cdots,k_{H}]$ and \textbf{V} = $[v_{1},v_{2},\cdot\cdot\cdot,v_{C}]$, receptively.\par
Next, we continue to use additive attention of vanilla Fastformer to summarize the query matrix into a global matrix $\textbf{q}\in \mathbb{R}^{1 \times HW\times 1}$, which aggregates global context information. In this process, $\alpha_{i}$ is calculated as follows:
\begin{equation}
\alpha_{i} = \frac{\text{exp}(\textbf{q}_{i})}{\sum_{i=1}^{H}\text{exp}(\textbf{q}_{i})}
\end{equation}
The global \textbf{q} matrix is calculated as follows:
\begin{equation}
\textbf{q} = \sum_{i=1}^{H}\alpha_{i}\times \textbf{q}_{i}
\end{equation}
Then we use element-wise product to realize the interactions between the global \textbf{q} matrix and every \textbf{k} matrix to obtain a global \textbf{p} matrix, which is calculated as $\textbf{p}_{i}=\textbf{q}\times \textbf{k}_{i}$. Similarly, $\beta_{i}$ is calculated as follows:
\begin{equation}
\beta_{i} = \frac{\text{exp}(\textbf{p}_{i})}{\sum_{i=1}^{H}\text{exp}(\textbf{p}_{i})}
\end{equation}
In vanilla Fastformer, the global \textbf{k} vector is obtained by element-wise product between $\beta$ and \textbf{p}. Unfortunately, such an approach leads to the loss of a large number of global features. Therefore, we propose using the matrix product (denoted as @) between $\beta$ and \textbf{p} to obtain the global \textbf{k} matrix. However, this creates a new problem: the computational cost becomes larger. Consequently, we use the average pooling(AP) to reduce the dimensions of $\beta$ and \textbf{p}. Finally, in order to maintain consistency in the size of every dimension, we perform bilinear interpolation(BI) of the resulting \textbf{k} matrix to obtain the final global \textbf{k} matrix. The entire calculation process is as follows:
\begin{equation}
\textbf{k} = \text{BI}(\sum_{i=1}^{H}(\text{AP}(\textbf{p}_{i})@\text{AP}(\beta_{i})))
\end{equation}
Finally, we still use matrix product to model the interaction between the global \textbf{k} matrix and every \textbf{v} matrix to obtain the \textbf{u} matrix, which is calculated as $\textbf{u}_{i}=\textbf{k}@\textbf{v}_{i}$. Then, the output is calculated as follows:
\begin{equation}
\text{VF}(\textbf{X}) = \text{Reshape}(\text{Concatenation}(\textbf{u}_{1},\textbf{u}_{2},\cdots,\textbf{u}_{C}))
\end{equation}
\begin{figure*}[t]
\centering
\includegraphics[width=0.6\textwidth]{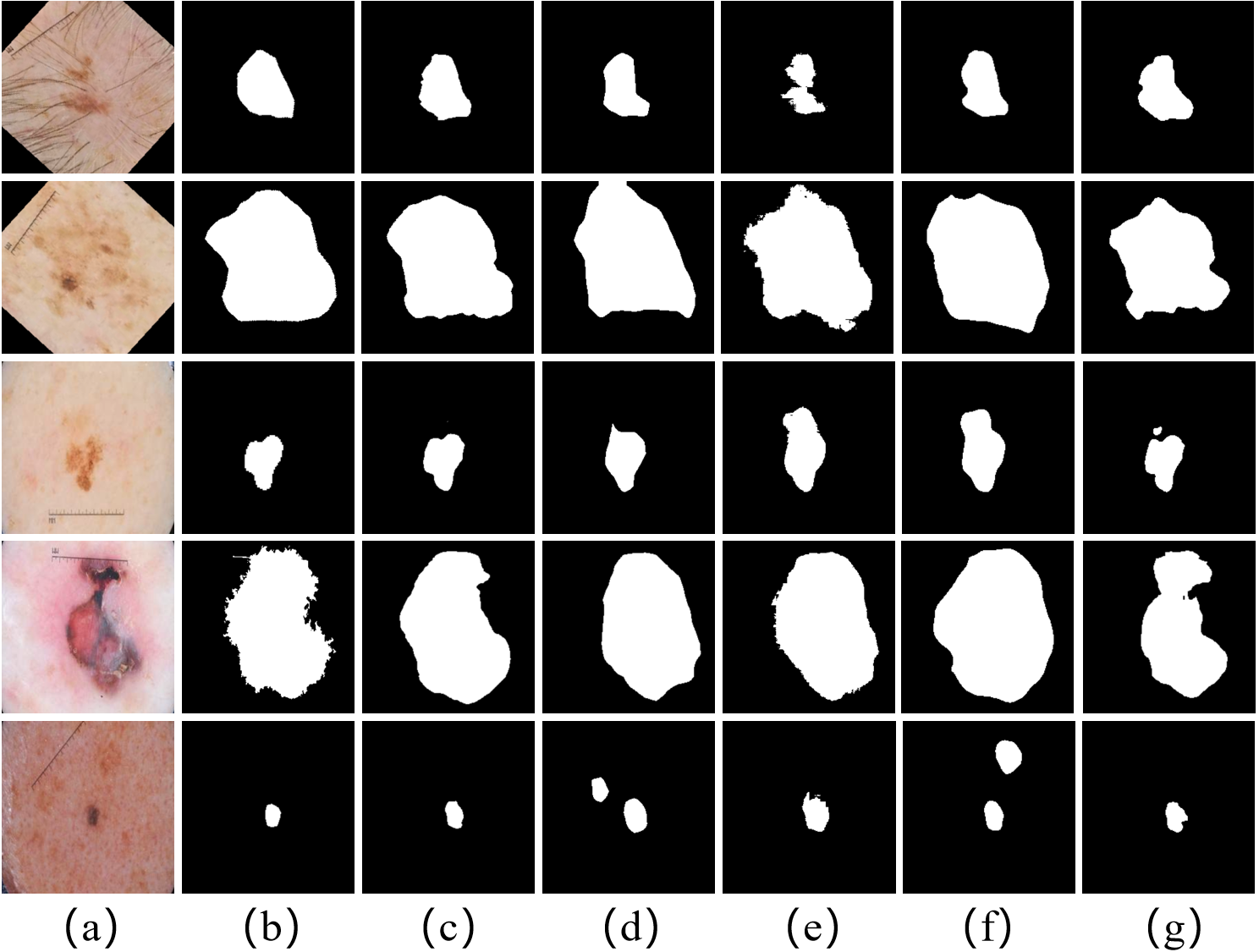}
\caption{Visual comparisons of different models on ISIC 2017. (a) Input images. (b) Groundtruth. The results by (c) VFFM-UNet, (d) ERDUNet, (e) C$^{2}$SDG, (f) MHorunet and (g) HSH-UNet.}
\label{fig121}
\end{figure*}
\begin{table*}[t]
\centering
\caption{Results for ISIC 2017 Dataset. VFFM-UNet shows significant advantages on segmentation performance.(\textbf{Bold} indicate the best and \uline{underline} indicate the second best.)}
\label{Tab2}
\begin{tabular}{l|c|cc|ccccc}
\hline
Methods & year & Params(M)↓ & GFLOPs↓ & mIoU↑ & DSC↑ & Acc↑ & Sen↑ & Spe↑ \\
\hline
UNet\cite{6} & 2015 & 31.04 & 54.74 & 77.08 & 87.06 & 95.82 & 84.68 & 96.93\\

UNet++\cite{37} & 2018 & 47.19 & 200.12 & 76.80 & 86.88 & 95.66 & 86.31 & 97.53\\

TransFuse\cite{7} & 2021 & 41.34 &  8.87 & 72.53 & 83.89 & 94.92 & 80.53 & \uline{97.99}\\

UTNet\cite{38} & 2021 & 15.29 & 22.55 & 76.79 & 86.87 & 95.63 & 86.76 & 97.41 \\

MISSFormer\cite{2} & 2022 & 35.45 & 7.28 & 76.97 & 86.98 & 95.81 & 84.14 & 97.94 \\

C$^{2}$SDG\cite{5} & 2023& 22.01 & 7.97 & 77.13 & 87.09 & 95.74 & 86.11 & 97.67\\

ERDUnet\cite{4} & 2024 & 10.21 & 10.29 & 78.19 & 87.76 & 95.96 & \uline{86.89} & 97.77 \\

MHorUNet\cite{1} & 2024& 3.49 & 0.57 & \uline{78.48} & \uline{87.94} & \uline{96.08} & 85.81 & 97.96\\

HSH-UNet\cite{3} & 2024 & 18.04 & 9.36 & 76.85 & 86.91 & 95.84 & 83.02 & 97.40\\
\hline
Ours & - & 0.35 & 0.494 & \textbf{79.00} & \textbf{88.32} & \textbf{96.31} & \textbf{87.11} & \textbf{98.22} \\
\hline
\end{tabular}
\end{table*}
\subsection{Vision Fastformer with Fusion Mechanism}
Since data with different degrees of severity are included in the sample lesion images, we introduce Vision Fastformer with Fusion Mechanism for feature maps to enhance the modelling capability of the model at multiple scales. As shown in Fig. \ref{fig100}, it contains three parts: Multi-Granularity Vision Fastformer, Multi-Granularity Fusion and Channel Fusion.\par
\textbf{Multi-Granularity Vision Fastformer}. First, We begin with Multi-Granularity Vision Fastformer, which consists of pixel-level VF(PiVF), patch-level VF(PaVF) and window-level VF(WiVF). The hierarchical structure can be used for local-neighbourhood modelling and global long-range dependency modelling. Pooling is a simple and effective method to enlarge the receptive field, which results in feature maps containing varying degrees of contextual information. Therefore, we use average pooling to progressively obtain feature maps at different granularities, including pixel-level, patch-level, and window-level. Given an input $\textbf{X}$, each feature map is computed as follows:
\begin{equation}
\begin{gathered}
\textbf{X}^{'} = \text{PiVF}(\textbf{X})=\text{VF}(\textbf{X}),\\
\textbf{X}^{''} = \text{PaVF}(\textbf{X}^{'})=\text{VF}(\text{AP}(\textbf{X}^{'})),\\
\textbf{X}^{'''} = \text{WiVF}(\textbf{X}^{''})=\text{VF}(\text{AP}(\textbf{X}^{''}))
\end{gathered}
\end{equation}
Where AP and VF denote average pooling and Vision Fastformer. Before entering fusion, we resize $\textbf{X}^{''}$ and $\textbf{X}^{'''}$ by bilinear interpolation(BI).
\begin{equation}
\textbf{X}^{''} = \text{BI}(\textbf{X}^{''}),\textbf{X}^{'''} = \text{BI}(\textbf{X}^{'''}),
\end{equation}
\textbf{Multi-Granularity Fusion(GF)}. After obtaining the feature maps with different granularities, we concatenate them and then pass through a 1×1 convolution layer $\mathcal{F}(\cdot)$ to achieve channel mixing:
\begin{equation}
\textbf{U} = \mathcal{F}(\text{Concatenate}(\textbf{X}^{'}, \textbf{X}^{''} \textbf{X}^{'''}))
\end{equation}
In order to efficiently extract the relationship between different granularities, we use channel-based average pooling and maximum pooling(MP):
\begin{equation}
\textbf{P} = \text{Concatenate}(\text{AP}(\textbf{U}), \text{MP}(\textbf{U}))
\end{equation}
Then, we apply a convolution layer $\mathcal{F}^{2\rightarrow3}$ followed by the sigmoid function $\sigma(\cdot)$ to obtain three weight masks: $\textbf{M}_{1}$, $\textbf{M}_{2}$ and $\textbf{M}_{3}$:
\begin{equation}
\textbf{M}_{1}, \textbf{M}_{2}, \textbf{M}_{3}=\sigma(\mathcal{F}^{2\rightarrow3}(\textbf{P}))
\end{equation}
Finally, we weight and sum the feature maps using the three weight masks and output(denoted as \textbf{Output of GF}) via a product residual connection:
\begin{equation}
\textbf{Output of GF}= (\textbf{X}^{'} \times \textbf{M}_{1} + \textbf{X}^{''} \times \textbf{M}_{2} + \textbf{X}^{'''} \times \textbf{M}_{3}) \times \textbf{X}
\end{equation}
\begin{figure*}[t]
\centering
\includegraphics[width=0.7\textwidth]{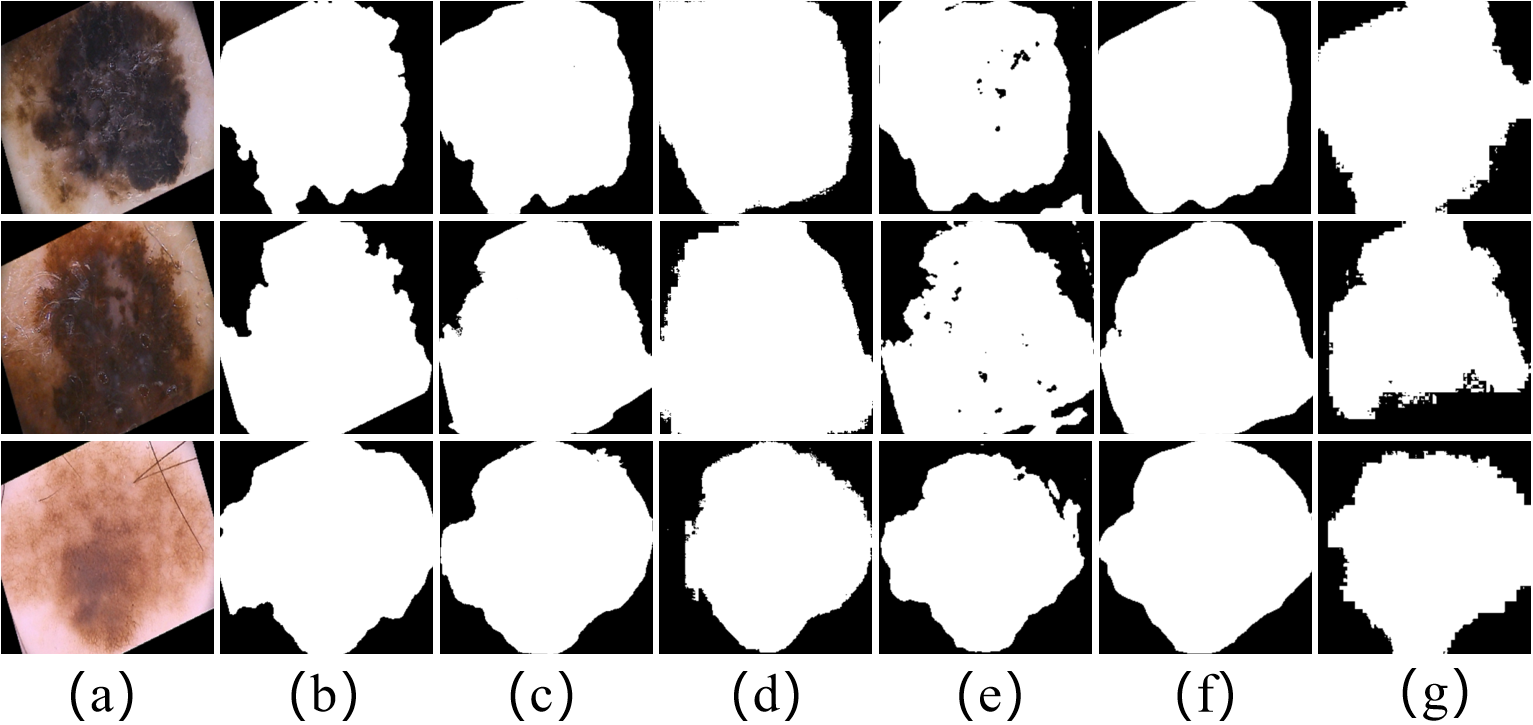}
\caption{Visual comparisons of different models on PH$^{2}$. (a) Input images. (b) Groundtruth. The results by (c) VFFM-UNet, (d) C$^{2}$SDG, (e) HSH-UNet, (f) MHorunet and (g)TransFuse.}
\label{fig6}
\end{figure*}
\begin{table*}[t]
\centering
\caption{Results for PH$^{2}$ Dataset. VFFM-UNet shows significant advantages on segmentation performance.(\textbf{Bold} indicate the best and \uline{underline} indicate the second best.)}
\label{Tab5}
\begin{tabular}{l|c|cc|ccccc}
\hline
Methods & year & Params(M)↓ & GFLOPs↓ & mIoU↑ & DSC↑ & Acc↑ & Sen↑ & Spe↑ \\
\hline
UNet\cite{6} & 2015 & 31.04 & 54.74 & 78.82 & 88.16 & 92.71 & 85.05 & \textbf{96.30}\\

UNet++\cite{37} & 2018 & 47.19 & 200.12 & 78.86 & 88.18 & 92.53 & 87.41 & 94.93 \\

TransFuse\cite{7} & 2021 & 41.34 &  8.87 & 76.44 & 86.65 & 91.18 & 89.72 & 91.87\\

UTNet\cite{38} & 2021 & 15.29 & 22.55 & 81.57 & 89.85 & 93.54 & 89.62 & 95.38 \\

C$^{2}$SDG\cite{5} & 2023& 22.01 & 7.97 & 81.63 & 89.88 & 93.46 & 91.08 & 94.58  \\

MHorUNet\cite{1} & 2024& 3.49 & 0.57 & 78.98 & 88.25 & 92.50 & 88.40 & 94.42 \\

HSH-UNet\cite{3} & 2024 & 18.04 & 9.36 & \uline{82.25} & \uline{90.26} & \uline{93.73} & \uline{91.14} & 94.14\\
\hline
Ours & - & 0.35 & 0.494 & \textbf{83.45} & \textbf{91.02} & \textbf{94.94} & \textbf{92.89} & \uline{95.85}\\
\hline
\end{tabular}
\end{table*}
\textbf{Channel fusion(CF)}. Processing the feature map in channel dimension helps the model achieve better segmentation performance. We introduce channel fusion to process feature maps at patch-level and window-level granularity.\par
First, we add $\textbf{X}^{''}$ and $\textbf{X}^{'''}$ together and then apply average pooling to the result:
\begin{equation}
\textbf{Y}=\text{AP}(\textbf{X}^{''} + \textbf{X}^{'''})=\frac{1}{H \times W}\sum_{i=1}^{H}\sum_{j=1}^{W}(\textbf{X}^{''}(i,j)+\textbf{X}^{'''}(i,j))
\end{equation}
Next, we perform the feature map $\textbf{Y}$ $\in $ $\mathbb{R}^{C\times H\times W}$ through a full convolution layer $\mathcal{F}^{C\rightarrow \frac{C}{2}}$ followed by batch normalization(BN) and the ReLU 
function $\delta(\cdot)$ to produce a new set of feature maps:
\begin{equation}
\textbf{Z}_{c}=\delta(\text{BN}(\mathcal{F}^{C\rightarrow \frac{C}{2}}(\textbf{Y})))
\end{equation}
We perform a full convolution layer $\mathcal{F}^{\frac{C}{2}\rightarrow C}$ again on the feature map $\textbf{Z}_{c}$ to adjust the channel number:
\begin{equation}
\textbf{Z}=\mathcal{F}^{\frac{C}{2}\rightarrow C}(\textbf{Z}_{c})
\end{equation}
Then, the feature map $\textbf{Z}$ is executed as the sigmoid function$\sigma(\cdot)$ to obtain two weight masks: $\textbf{W}_{1}$ and $\textbf{W}_{2}$:
\begin{equation}
\textbf{W}_{1}=\sigma(\textbf{Z}),\textbf{W}_{2}= 1-\textbf{W}_{1}
\end{equation}
Finally, we weigh and sum the feature maps at patch-level and window-level granularity. The output(denoted as \textbf{Output of CF}) is calculated as follows:
\begin{equation}
\textbf{Output of CF}=\textbf{X}^{''} \times \textbf{W}_{1} + \textbf{X}^{'''} \times \textbf{W}_{2}
\end{equation}
\section{Experiments}
In this section, we first introduce the datasets and the implementation details. Then we compare our
experimental results with several of the most popular medical image segmentation models and general-purpose models. In Fig. \ref{fig7}, notably, VFMFM-UNet achieves state-of-the-art in terms of an optimal balance between the number of parameters, computational complexity, and segmentation performances. Finally, we will conduct ablation studies to validate the effectiveness of our proposed modules. In addition, we further explore the effect of head number on the model's performance to determine this important hyperparameter.
\subsection{Datasets}
To validate the effectiveness of our model, we conduct extensive comparisons with state-of-the-art models on three public lesion segmentation datasets: the International Skin Imaging Collaboration 2017 and 2018 challenge datasets (ISIC2017 and ISIC2018) and PH$^{2}$ datasets.\par
The ISIC2017 dataset contains 2,150 dermoscopic images with corresponding segmentation mask labels. We follow the same data processing approach for this dataset as in prior research. The dataset is first divided into training, validating and testing subsets using a 7:3 ratio. Specifically, 1,500 images are allocated for training, and 650 images are reserved for validating and testing. \par
The ISIC2018 dataset contains 2,694 dermoscopic images with corresponding segmentation mask labels. Following the methodology described in ISIC2017, the dataset is divided into training, validating and testing subsets. Specifically, 1,886 images are used for the training set and 808 images are reserved for validating and testing. \par
A total of 200 challenging images were collected from PH$^{2}$ dataset, along with dermoscopic images including segmentation mask labels. Specifically, we loaded the weights trained on ISIC2017 dataset and applied them to test on this dataset, further demonstrating the model's generalization capability.
\subsection{Implementation Details}
Our VFFM-UNet is implemented on PyTorch 2.0.0. All the experiments are conducted on a single NVIDIA RTX 4060 GPU. The input images are uniformly normalized and resized to 256 × 256 in our preprocessing process. Additionally, we apply data augmentation techniques such as vertical flipping, horizontal flipping, and random rotations. AdamW is the optimizer set with an initial learning rate of 1e-4 and weight decay of 1e-2. At the same time, the CosineAnnealingLR is adopted as the learning rate scheduler with a maximum of 50 iterations and a minimum learning rate of 1e-5. We set different epochs for different datasets: 220 for ISIC2018 and 240 for ISIC2017. For training, we set the batch size to 8 and utilize a combined loss function that includes both the Dice loss $\text{L}_{Dice}$ and the cross-entropy loss $\text{L}_{CE}$, defined as follows:
\begin{equation}
\text{L}_{all} = \omega \text{L}_{Dice} + (1 - \omega) \text{L}_{CE}
\end{equation}
where $\omega = 0.6$ and $1 - \omega = 0.4$ are receptively weights for the Dice loss and the cross-entropy loss.
\begin{figure}[t]
\centering
\includegraphics[width=0.35\textwidth]{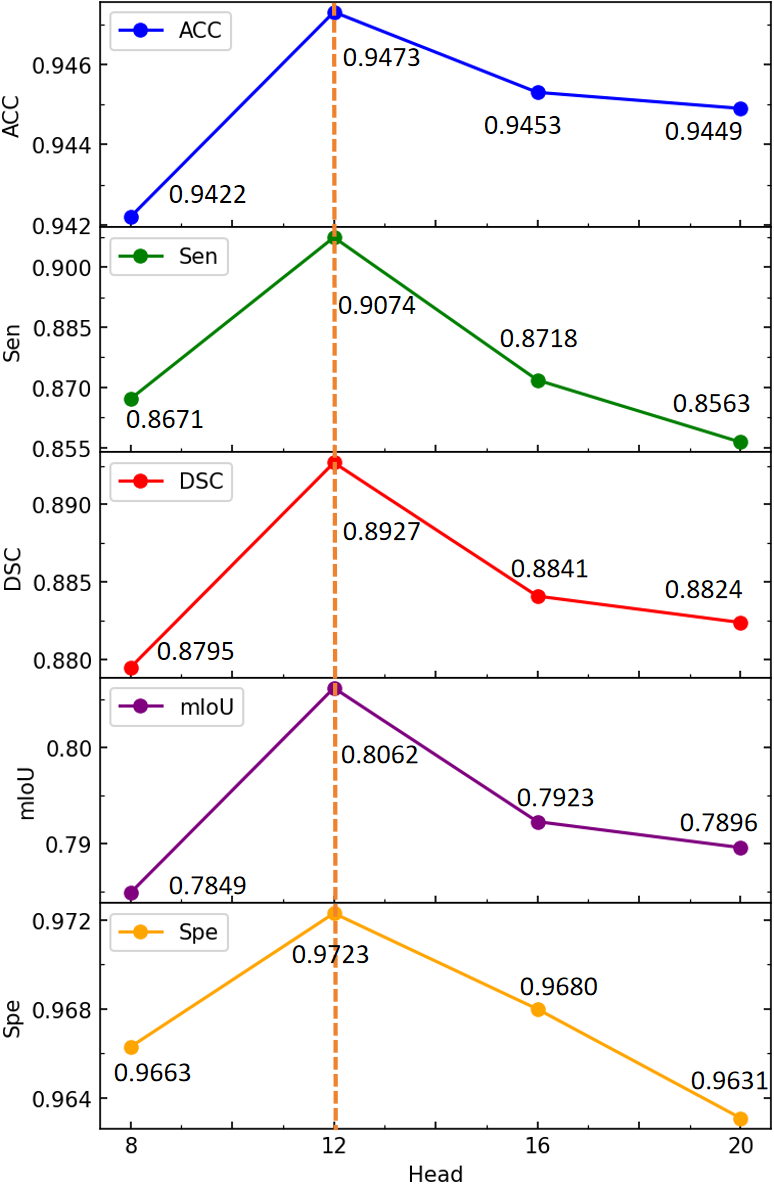}
\caption{Performance of VFFM-UNet in different head numbers}
\label{fig8}
\end{figure}
\subsection{Evaluation Metrics}
We employed five different metrics to assess the performance of the segmentation: Mean Intersection over Union (mIoU), Dice Similarity Score (DSC), Accuracy (Acc), Sensitivity (Sen) and Specificity (Spe). The mathematical definitions of these metrics are outlined as follows:
\begin{equation}
\text{mIoU}=\frac{\text{TP}}{\text{TP}+\text{FP}+\text{FN}}
\end{equation}
\begin{equation}\text{DSC}=\frac{\text{2TP}}{\text{2TP}+\text{FP}+\text{FN}}
\end{equation}
\begin{equation}
\text{Acc}=\frac{\text{TP}+\text{TN}}{\text{TP}+\text{TN}+\text{FP}+\text{FN}}
\end{equation}
\begin{equation}
\text{Sen}=\frac{\text{TP}}{\text{TP}+\text{FN}}
\end{equation}
\begin{equation}
\text{Spe}=\frac{\text{TN}}{\text{TN}+\text{FP}}
\end{equation}
Where \text{TP}, \text{FP}, \text{FN}, \text{TN} represent true positive, false positive, false negative, and true negative.
\subsection{Comparison results}
To emphasize the performance of our model, we use the five different evaluation metrics to compare the experimental results of VFFM-UNet with other current advanced models, including UNet++\cite{37}, TransFuse\cite{7}, UNet\cite{6}, UTNet\cite{38}, C$^{2}$SDG\cite{5}, ERDUnet\cite{4}, MISSFormer\cite{2}, HSH-UNet\cite{3}, and MHorUNet\cite{1}. At the same time, we calculate the number of parameters and GFLOPs for each model to evaluate computational costs. Notably, for fairness, We reimplement these above models with the same computing environments and hyperparameter settings according to the publicly released codes. This demonstrates that the model's performance improvement is due to the changes in the model architecture rather than adjustments to hyperparameter settings.
\subsubsection{Results on ISIC 2018 Dataset}
On the ISIC 2018 dataset, our VFFM-UNet outperforms several state-of-the-art methods. The quantitative results are shown in Tab. \ref{Tab3}. Specifically, compared to large models like MISSFormer, our model not only exceeds their performance but also reduces the number of parameters and computations by a factor of 101x and 15x, receptively. Compared to the lightweight model, VFFM-UNet achieves increases of about 1.44\%, 0.89\% and 3.9\% more than MHorUNet in mIoU, DSC and Sen metrics. To further demonstrate the advantages of our model, we select some challenging examples from the ISIC 2018 dataset for visualization, which are generated by C$^{2}$SDG\cite{5}, ERDUnet\cite{4}, MISSFormer\cite{2}, HSH-UNet\cite{3}, and MHorUNet\cite{1}.\par
As shown in Fig. \ref{fig4}, these challenging images generally have the following characteristics: some images have small, prominent lesions, where the lesion's boundaries are unclear and indistinct due to variations in skin colour. Others show more obvious lesion areas, which are large but not clumped together, presenting as irregularly sized spots, making it difficult to distinguish the actual lesion boundaries. It can be observed that VFFM-UNet has generally achieved satisfactory results, but other five methods encounter issues when processing the above data, such as rough segmentation boundaries (e.g., ERDUNet in the 4th and 5th rows of Fig. \ref{fig4}), significant loss of lesion areas (e.g., MHorunet, C$^{2}$SDG and HSH-UNet in the 1st and 2nd rows of Fig. \ref{fig4}) and over-segmentation (e.g., MHorunet in the 1st row and C$^{2}$SDG in the 5th row of Fig. \ref{fig4}).\par

\begin{figure}[t]
\centering
\includegraphics[width=0.3\textwidth]{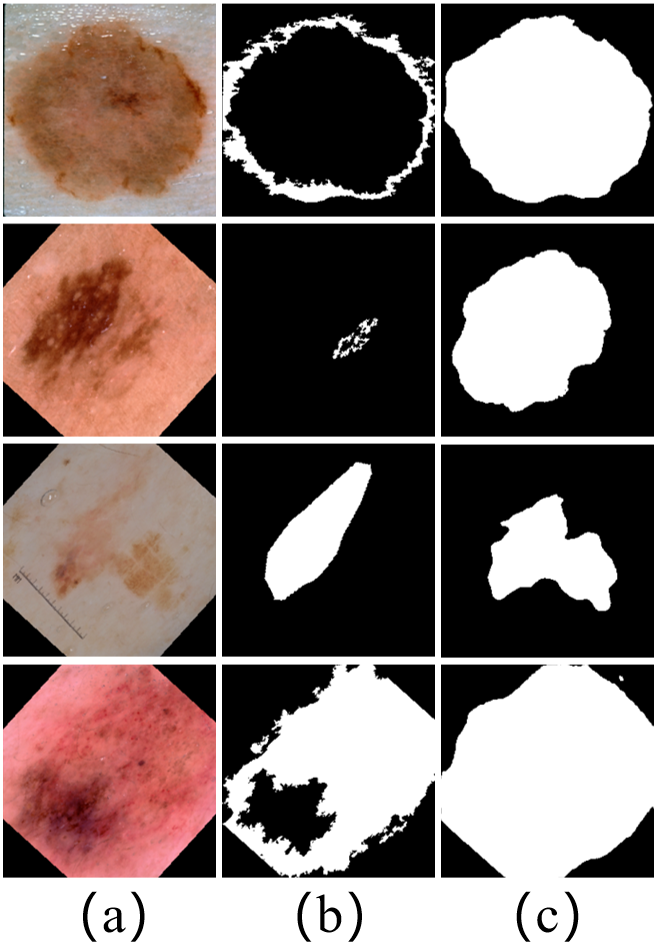}
\caption{Visualization of some mislabeled data. (a) Input images. (b) Groundtruth. (c) Results of our model.}
\label{fig5}
\end{figure}

It should be noted that there are some mislabeled data in the testing dataset. Since our data preprocessing follows the method from previous articles, we did not exclude such data in advance. As illustrated in Fig. \ref{fig5}, our model still produces satisfactory segmentation results.
\subsubsection{Results on ISIC 2017 Dataset}
We evaluate our VFFM-UNet on the ISIC 2017 dataset, as shown in Tab. \ref{Tab2}. Taking a model as an example, while achieving lightweight design, our model also shows significant increases of about 0.81\%, 0.56\%, 0.35\%, 0.22\%, and 0.45\% more than ERDUnet\cite{4} in mIoU, DSC, Acc, Sen, and Spe metrics. As shown in Fig. \ref{fig121}, thanks to the Fastformer module, our model can model long-range dependencies, allowing it to handle lesion boundaries with a low severity from a broader perspective(e.g., the visual comparisons in the first four rows.). In addition, for regular spots on the skin, our model is less influenced by them and can segment the lesions more accurately while ensuring no erroneous inferences in these regions(e.g., the visual comparisons in the last rows.).
\subsubsection{Results on PH$^{2}$ Dataset}
To further verify our VFFM-UNet, we conduct experiments
on PH$^{2}$. Unlike the previous three datasets, which feature large-scale data distribution and unclear lesion
boundaries, the PH$^{2}$ dataset consists of only a few hundred dermoscopic images, with the lesions being severe, resulting in a clearer contrast between the lesions and normal skin. The results are listed in Tab. \ref{Tab5}. This indicates that our model demonstrates good performance, which are 83.45\%, 91.02\%, 94.94\%, 92.89\% and 94.85\% in mIoU, DSC, Acc, Sen, and Spe metrics. The results emphasize the strong generalization capability of our model. As shown in Fig. \ref{fig6}, our VFFM-UNet can also extract more detailed structural information and produce more precise edges when processing images with high contrast. Such results are attributed to the fusion mechanism. This mechanism allows our model to obtain more comprehensive feature maps from both the granularity and channel dimensions, enabling it to process lesion images with different degrees of severity dynamically.
\subsection{Ablation results}
We conduct comprehensive ablation experiments on the ISIC2017 dataset to validate the effectiveness of our proposed modules. These experiments involve assessing the performance of the VF and FM components. The baseline utilized in our work is a six-stage U-shaped architecture with symmetric encoder and decoder parts and a plain skip connection. Each stage includes a plain convolution operation with a kernel size of 3, and the number of channels is set to \{8, 16, 24, 32, 48, 64\}. In addition, Average pooling is used for downsampling.\par
\subsubsection{Effects of Vision Fastformer}We add Vision Fastformer module in the last three stages of the baseline. As shown in the Tab. \ref{Tab1}, VF improves the performance of the model. This demonstrates that capturing global contextual information to enable long-range dependency modelling is critical and indispensable.\par
\subsubsection{Effects of Fusion Mechanism}To validate the effectiveness of obtaining different levels of contextual information on segmentation performance, we add Fusion Mechanism to VF module. As shown in Tab. \ref{Tab1}, FM contributes to the model's performance. This suggests that feature maps with different granularities and the processing of feature maps in the channel dimension effectively guide the model's inferences.
\subsubsection{All you need is more than just the element-wise product}
In the third row of Tab. \ref{Tab1}, we present the quantitative results obtained using the vanilla Fastformer (denoted as VF$_{base}$ in the table, which refers to Fastformer without matrix product and channel reduction). It can be observed that, compared to the improved module, Vision Fastformer, there is a significant gap in the mIoU, DSC, Acc, Sen, and Spe metrics, with decreases of 1.64\%, 1.07\%, 0.12\%, 2.17\%, and 0.21\%, respectively. Through a detailed analysis of the underlying issue, we can draw the following insightful conclusion: Image data is inherently two-dimensional, with strong pixel correlations. The element-wise product operates only on the pixels at corresponding positions in feature maps, lacking interaction between pixels and, consequently, interaction between different features. This neglects the context and results in poor performance in capturing global contextual information. Therefore, we combine the element-wise product and matrix product for improvement, aiming to retain more feature information through linear combinations, thereby better capturing the global structure of the data.
\subsubsection{Is the head number of VFFM-UNet Important?}
In Fastformer module, the number of heads of self-attention is a significant hyperparameter to learn global contextual information. Considering the balance between computational costs and long-range dependency modelling, the number of heads we have selected in the experiments is set to \{8, 12, 16, 20\}. Fig. \ref{fig8} shows the result in ISIC2018 in different numbers of heads. Consequently, we can draw the following conclusions. The five metrics exhibit a trend of increasing and then decreasing, which is maximized at 12. When the number of heads is too small, the extraction of feature maps is very rough and imprecise. With an increasing number of heads, the model gains more perspectives in handling global contextual information. However, errors between the model's inference results and the actual values still exist and accumulate with this increase. Therefore, a considerable number of heads results in reduced performance. Consequently, we choose 12 as the best head number in our model.
\begin{table}[t]
\centering
\setlength{\tabcolsep}{3pt} 
\caption{Ablation studies on the ISIC2017 dataset.}
\label{Tab1}
\begin{tabular}{l|llllll} 
\hline
Methods & mIoU↑ & DSC↑ & Acc↑ & Sen↑ & Spe↑\\
\hline
baseline & 76.71 & 86.82 & 95.67 & 85.74 & 97.65 \\
baseline + VF & 78.40 & 87.89 & 95.81 & 86.29 & 97.88 \\
baseline + VF$_{base}$ + FM & 77.36 & 87.25 & 95.90 & 84.94 & 98.01 \\
baseline + VF + FM & 79.00 & 88.32 & 96.02 & 87.11 & 98.22 \\
\hline
\end{tabular}
\end{table}
\section{Discussion and Conclusion}
To identify unclear lesion boundaries, especially in samples with subtle colour changes, We explored the potential of FastFormer in medical segmentation and integrated this module into the UNet architecture. Our Vision Fastformer follows an additive attention mechanism to summarize the query and key matrix into a global matrix and combines element-wise product and matrix product to optimize the balance between computational costs and long-range dependency modelling. To achieve generalization on lesion boundaries of different severity, we proposed Fusion Mechanism. This module processes and fuses the feature maps extracted by Vision FastFormer in both the granularity and channel dimensions, enabling the model to have various perspectives and dynamically adjust the required contextual information for different lesion images. Quantitative and qualitative analyses demonstrate that VFFM-UNet sets a new benchmark by achieving an optimal balance between parameter numbers, computational complexity, and segmentation performance compared to existing state-of-the-art models.\par
\begin{figure}[t]
\centering
\includegraphics[width=0.3\textwidth]{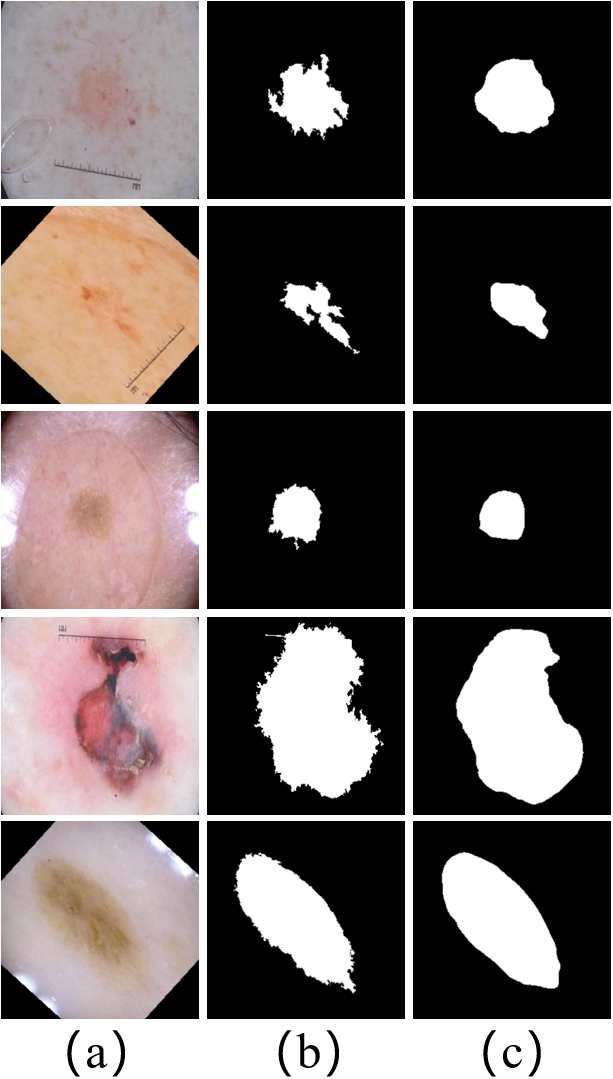}
\caption{Visualization of some failure datas. (a) Input images. (b) Groundtruth. (c) Results of our model.}
\label{fig11}
\end{figure}

Our model also has some limitations in handling some
extremely challenging cases. As shown in Fig. \ref{fig11}, our model's segmentation ability still has room for improvement when handling skin lesion images with rough, irregular edges rather than smooth, curve-like boundaries. Besides, our model is only used for skin lesion segmentation, and our research will explore the application of Fastformer to other medical image segmentation in the future.

\section{Acknowledgments}
This work was supported by the Natural Science Foundation of Shandong Province, China (Grant no. ZR2015FM014) and the National Natural Science Foundation of China (grant Nos. 42192535 \& 42274006 \& 42104084).

\section{Conflict of Interest Statement}
The authors declare no conflicts of interest.

\section*{References}
\addcontentsline{toc}{section}{\numberline{}References}

\bibliographystyle{medphy.bst} 

\end{document}